\documentclass[10pt,twocolumn,letterpaper]{article}
\usepackage{authblk}

\usepackage[pagenumbers]{cvpr} 

\usepackage[ruled,linesnumbered]{algorithm2e}
\usepackage{multirow}
\usepackage{bbding}
\usepackage{makecell}

\definecolor{cvprblue}{rgb}{0.21,0.49,0.74}
\usepackage[pagebackref,breaklinks,colorlinks,citecolor=cvprblue]{hyperref}

\def\paperID{4273} 
\def\confName{CVPR}
\def\confYear{2024}

\title{ActiveDC: Distribution Calibration for Active Finetuning}

\author[1, 2]{Wenshuai Xu}
\author[2]{Zhenghui Hu \thanks{Corresponding author}}
\author[2]{Yu Lu}
\author[1, 2]{Jinzhou Meng}
\author[2, 3]{Qingjie Liu}
\author[2, 3]{Yunhong Wang}
\affil[1]{School of Software, Beihang University}
\affil[2]{Hangzhou Innovation Institute, Beihang University}
\affil[3]{State Key Laboratory of Virtual Reality Technology and Systems, Beihang University}

\begin{document}
\maketitle
\begin{abstract}
The pretraining-finetuning paradigm has gained popularity in various computer vision tasks. 
In this paradigm, the emergence of active finetuning arises due to the abundance of large-scale data and costly annotation requirements. 
Active finetuning involves selecting a subset of data from an unlabeled pool for annotation, facilitating subsequent finetuning. 
However, the use of a limited number of training samples can lead to a biased distribution, potentially resulting in model overfitting. 
In this paper, we propose a new method called ActiveDC for the active finetuning tasks. 
Firstly, we select samples for annotation by optimizing the distribution similarity between the subset to be selected and the entire unlabeled pool in continuous space. 
Secondly, we calibrate the distribution of the selected samples by exploiting implicit category information in the unlabeled pool. 
The feature visualization provides an intuitive sense of the effectiveness of our method to distribution calibration. 
We conducted extensive experiments on three image classification datasets with different sampling ratios. 
The results indicate that ActiveDC consistently outperforms the baseline performance in all image classification tasks. 
The improvement is particularly significant when the sampling ratio is low, with performance gains of up to 10\%. 
Our code will be publicly available.
\end{abstract}

\begin{figure}[!t]
\centering
\includegraphics[width=1.0\linewidth]{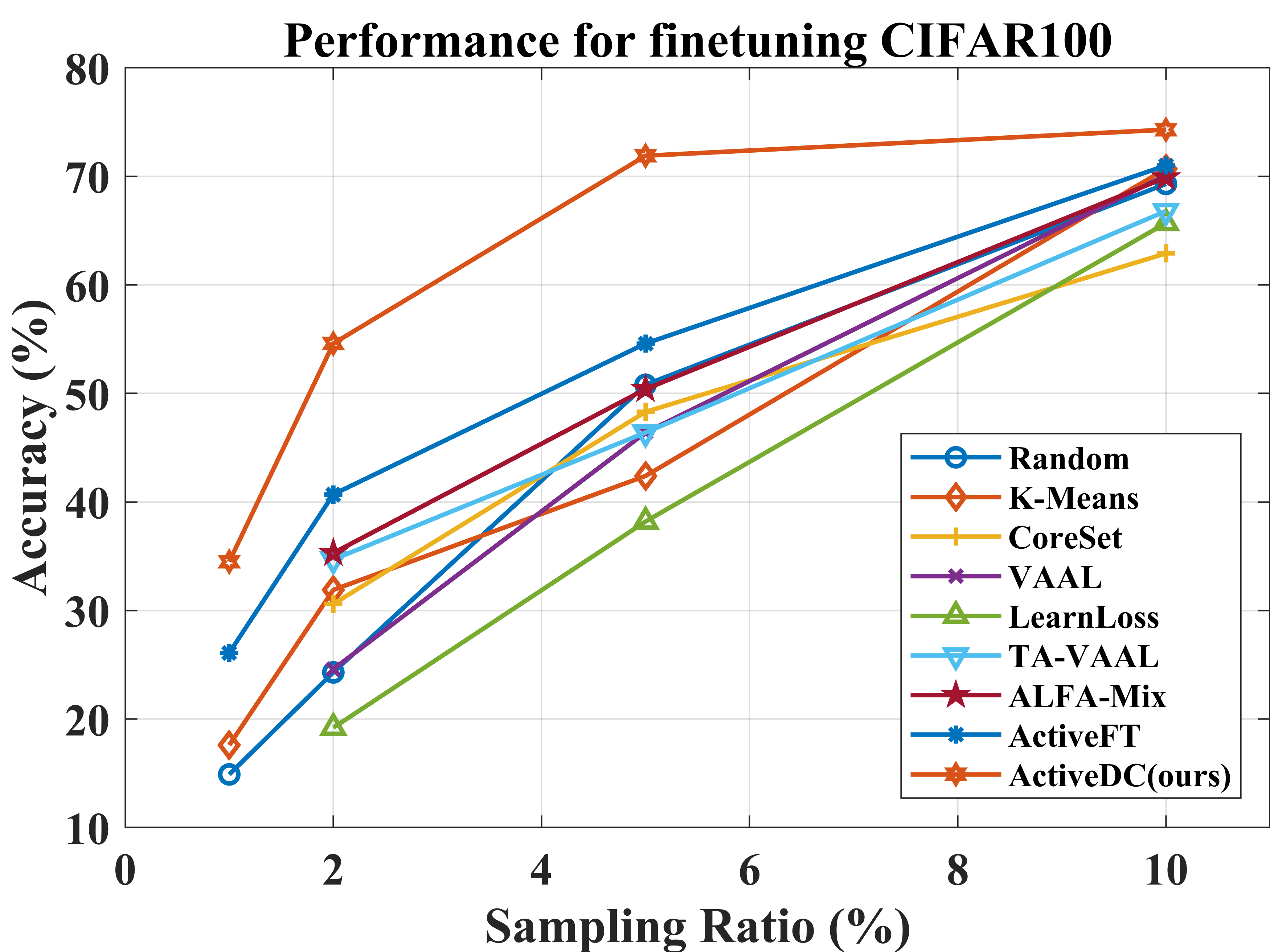}
\caption{Comparison of the performance for finetuning the CIFAR100 dataset at different sampling ratios.
}
\label{fig:comp_perf_cifar100}
\end{figure}

\begin{figure*}[!t]
\centering
\includegraphics[width=1.00\linewidth]{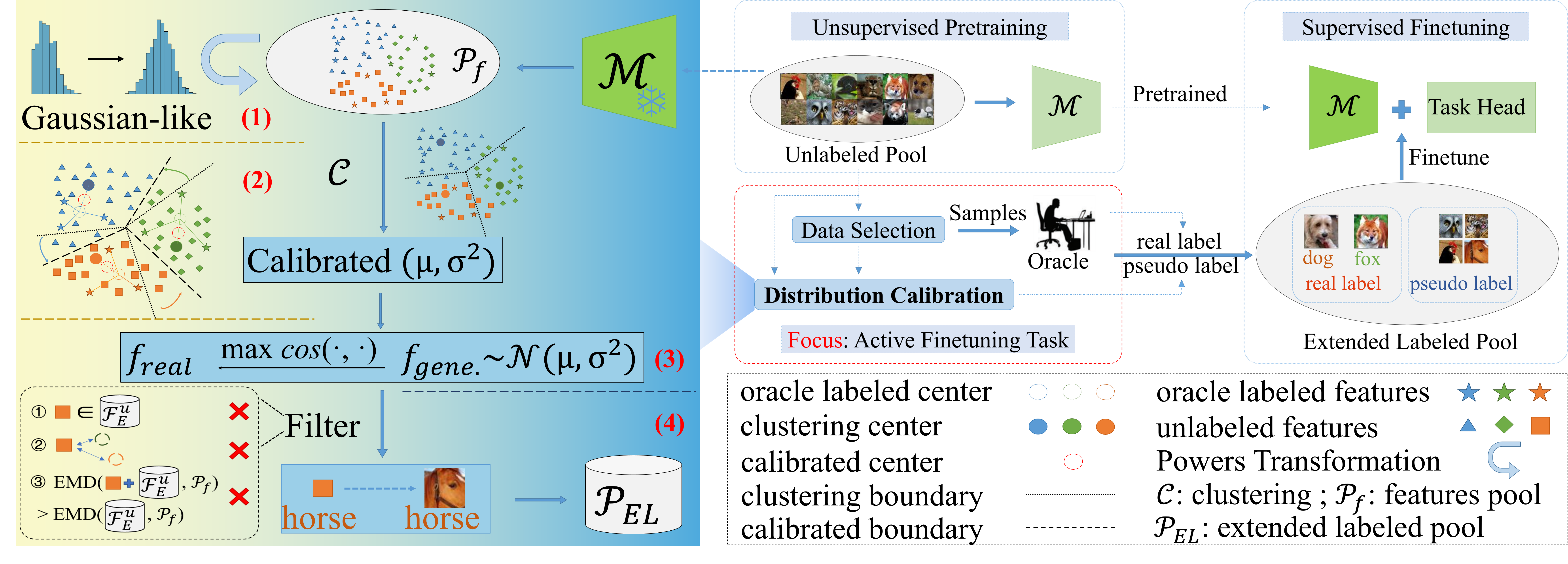}
\caption{
The active finetuning task involves the active selection of training data for finetuning within the pretraining-finetuning paradigm.
We focus on data selection and distribution calibration from a large unlabeled data pool for annotation.
The Distributed Calibration Module comprises four main steps: (1) applying Tukey's Ladder of Powers Transformation to render the feature distribution more Gaussian-like, (2) clustering the features and calibrating the statistics for different feature classes, (3) generating pseudo-features using the calibrated statistics and identifying the most similar real features, and (4) filtering and integrating the features into the extended labeled pool.
}
\label{fig:pretrainng_finetuning}
\end{figure*}

\section{Introduction}
\label{sec:intro}

The recent successes in deep learning owe much of their progress to the availability of extensive training data.
However, it is crucial to recognize that annotating large-scale datasets demands a significant allocation of human resources.
In response to this challenge, a prevalent approach has emerged, referred to as the pretraining-finetuning paradigm.
This paradigm involves the initial pretraining of models on a substantial volume of data in an unsupervised fashion, followed by a subsequent finetuning phase on a more limited, labeled subset of data.

The existing body of literature extensively delves into the domains of unsupervised pretraining~\cite{SimCLR_ICML2020,MoCo_CVPR2020,iBOT_ICLR2022} and supervised finetuning~\cite{liu2023pre,ShrinkMatch_ICCV2023}, making noteworthy contributions in these areas. 
However, when confronted with a vast repository of unlabeled data, the crucial task at hand is the judicious selection of the most valuable samples for annotation, a task necessitated by the constraints of limited annotation resources. 
At the same time, the distribution of the small number of labeled samples tends to significantly deviate from the overall distribution, raising the issue of how to calibrate the distribution of the selected samples~\cite{FreeLunch_ICLR2021,ALB_ICLR2021}.

Active learning~\cite{CoreSet_ICLR2018,BGADL_ICML2019,BADGE_ICLR2020,Box_Level_AD_CVPR2023,HAL_clustering_VAD_CVPR2023,Active_SSL_ICCV2023,ReFAL_CVPR2023} is a method that iteratively selects the most informative samples for manual labeling during the training process, aiming to improve predictive model performance. 
Although active learning is considered a promising approach, empirical experiments have uncovered its limitations~\cite{SSmeetAL_ICCV2021,ActiveFT_CVPR2023,TypiClust_ICML2022} when employed within the context of the pretraining-finetuning paradigm.
One plausible explanation for this phenomenon is the presence of more constrained annotation budgets and batch-selection strategy of active learning, which introduce biases into the training process, ultimately impeding its effectiveness.

In response to the limitations of traditional active learning within the pretraining-finetuning paradigm, a more efficient active finetuning approach, known as ActiveFT~\cite{ActiveFT_CVPR2023}, has been developed to address these challenges.
This method selects sample data by narrowing the distribution gap between the chosen subset and the entire unlabeled pool.
Although the method exhibits promising results, its primary focus is distributional information. 
Unfortunately, it does not sufficiently harness information related to the number of known classification categories, and the implicit category-related data within a substantial volume of unlabeled pretrained features remains underutilized. 
Importantly, when working with a limited number of selected samples, there is \emph{a heightened risk of bias} in how the chosen subset aligns with the overall distribution~\cite{FreeLunch_ICLR2021,ALB_ICLR2021}.
Which, in turn, necessitates a larger sample size to rectify the distributional alignment.

In this paper, we present a novel method, \textbf{ActiveDC}, designed for enhancing active finetuning tasks.
We introduce a novel distributional calibration technique to the active finetuning task. 
Our method leverages the comprehensive information derived from the entire feature data pool along with the labeled data, resulting in a significant improvement in model performance. 
Importantly, it achieves this enhancement while remaining cost-effective, eliminating the need for additional labeling efforts. 
Moreover, it does not incur excessive time consumption.

Specifically, our method comprises several key steps, as shown in \cref{fig:pretrainng_finetuning}. 
First, we employ a diversity selection strategy, such as ActiveFT, to select data for oracle annotation within limited budgets. 
Second, we proceed to normalize and transform the feature distribution of the entire data pool extracted by the pretrained model, followed by the application of clustering techniques. The resulting clustering categories are determined in reference to the labeled data, taking into account a trade-off between the center of clustering and the center of labeled samples, as dictated by the quantity of labeled samples available. Furthermore, we actively regulate the covariance of pseudo-categorical features. Subsequently, pseudo-features are generated based on the calibrated statistics, and real features most similar to these pseudo-features are identified through a similarity-based approach. The corresponding real data is then pseudo-labeled and integrated into the expanded labeled pool for finetuning. Throughout the iterative process of feature generation, certain filtering procedures are applied, contingent on the influence of the generated data on the extended labeled pool.
Notably, the method excels in calibrating the distribution of a limited subset of training data selected by the active finetuning procedure, as depicted in \cref{fig:comp_perf_cifar100}.

Our main contributions are summarized as follows:
\begin{itemize}
    \item We propose a new method, ActiveDC, aimed at calibrating data distributions for samples chosen through active finetuning techniques within limited labeling budgets. 
    Our method significantly improves classification accuracy while maintaining the same budget.
    \item The Distributed Calibration Module exhibits inherent flexibility, allowing seamless integration with various active finetuning selection strategies.
    \item In the context of the classification task, our method yields significant performance improvements, with particular emphasis on its effectiveness at lower sampling ratios.
\end{itemize}

\section{Related Work}
\label{sec:work}

\textbf{Unsupervised Learning} is designed to acquire feature representations without the need for labeled data. 
It plays a crucial role within the pretraining-finetuning paradigm. Besides, it exhibits notable effectiveness in mitigating the laborious task of data labeling~\cite{SSmeetAL_ICCV2021}.
Both contrastive methods~\cite{SimCLR_ICML2020,MoCo_CVPR2020,SwAV_NIPS2020,BYOL_NIPS2020,SimSiam_CVPR2021} and generative methods~\cite{MAE_CVPR2022,BeiT_ICCV2022,SparseMAE_ICCV2023,PerceptualMAE_CVPR2023,singh2023effectiveness} have achieved remarkable success in this field.
The core idea behind contrastive learning is to train a model to distinguish between similar and dissimilar pairs of data samples.
Building on the successes of existing contrastive learning techniques and the remarkable achievements of vision transformers in computer vision, innovative approaches like MoCov3~\cite{MoCov3_ICCV2021}, DINO~\cite{DINO_ICCV2021}, and iBOT~\cite{iBOT_ICLR2022} have successfully extended the principles of contrastive learning to the realm of vision transformers~\cite{ViT_ICLR2021}, further enriching this vibrant field of study.
Recent research in generative methods~\cite{MAE_CVPR2022,BeiT_ICCV2022} for predicting masked content within input samples has demonstrated promising performance over vision transformers.
Extensive prior research~\cite{ActiveFT_CVPR2023,SSmeetAL_ICCV2021,TypiClust_ICML2022,SparseMAE_ICCV2023,PerceptualMAE_CVPR2023} has thoroughly examined the advantageous contributions of both kinds of methods in the context of downstream supervised finetuning.

\textbf{Active Learning} aims at selecting the most valuable samples for labeling to optimize the model performance with a limited labeling budget.  
Much of the existing research in this field centers on sample selection strategies based on uncertainty~\cite{BvSB_CVPR2009,LLoss_CVPR2019,VAAL_ICCV2019,TA-VAAL_CVPR2021,ALFA-Mix_CVPR2022} and diversity~\cite{BADGE_ICLR2020,CoreSet_ICLR2018,WAAL_ICAIS2020,TypiClust_ICML2022} in pool-based scenarios. 
Uncertainty quantifies the model's level of perplexity when presented with data and can be estimated through various heuristics, including predictive probability, entropy, margin~\cite{BvSB_CVPR2009} and predictive loss~\cite{LLoss_CVPR2019}. 
Conversely, some algorithms seek to identify a subset that effectively represents the entire data pool by considering the diversity and representativeness of the data~\cite{CSVAL_MIDL2023}.
Diversity is quantified using measures such as Euclidean distance between global features~\cite{CoreSet_ICLR2018}, KL-divergence~\cite{CDAL_ECCV2020} between local representations, gradients spanning diverse directions~\cite{BADGE_ICLR2020}, or adversarial loss~\cite{VAAL_ICCV2019,TA-VAAL_CVPR2021,WAAL_ICAIS2020}, and more. 

\textbf{Active Finetuning} is a task that actively selects training data for finetuning within the pretraining-finetuning paradigm.
The majority of active learning algorithms mentioned above are primarily tailored for training models from scratch. 
However, prior research~\cite{SSmeetAL_ICCV2021,ActiveFT_CVPR2023,TypiClust_ICML2022}, has elucidated their adverse effects when applied to the finetuning process following unsupervised pretraining. 
Therefore, the introduction of the active finetuning task presents a novel approach to sample selection for labeling in a single pass~\cite{ActiveFT_CVPR2023,FreeSel_NIPS2023}, especially in preparation for subsequent finetuning processes. 
An effective strategy, named ActiveFT~\cite{ActiveFT_CVPR2023}, involves selecting data for labeling by converging the distribution of the chosen subset with that of the complete unlabeled pool within a continuous space.
Nevertheless, unsupervised models, derived from extensive pretraining on large-scale datasets, exhibit robust feature extraction capabilities. As a result, within the pretraining-finetuning paradigm, the selection of samples often involves a limited quantity. 
In such cases, the inherent bias in the subset chosen by ActiveFT to align with the overall distribution tends to be relatively pronounced, necessitating a greater volume of samples to rectify the distribution effectively.

\section{Methodology}
\label{sec:method}

This section delineates our novel active finetuning methodology. 
\Cref{sec:method:overview} initiates by furnishing a comprehensive overview of the pretraining-finetuning paradigm, followed by \cref{sec:method:dist_sele}, which introduces the data selection module. 
Subsequently, \cref{sec:method:dist_cali} furnishes an exhaustive exposition of the distribution calibration module.

\subsection{Overview}
\label{sec:method:overview}
The complete pipeline for conducting the active finetuning task within the pretraining-finetuning paradigm is visually represented in \cref{fig:pretrainng_finetuning}. 
This paradigm consists of two distinct stages. In the initial stage, the model undergoes unsupervised pretraining on an extensive dataset, enabling it to traverse various classes of data features within the feature space. This stage establishes the foundation for subsequent feature extraction. In the next stage, the pretrained model is coupled with a task-specific module to facilitate supervised finetuning on a smaller, labeled subset tailored to specific tasks.
The pivotal juncture between these two stages centers on the meticulous construction of the labeled subset. 
We select data for labeling based on fitting the distribution of the entire feature pool, and select appropriate data for pseudo-labeling by calibrating the category distributions to construct well-distributed training data for subsequent finetuning.

We formally define a deep neural network model $\mathcal{M}(\cdot;\omega_0) : \mathcal{X} \xrightarrow{} \mathbb{F}^u$ with pretrained weight $\omega_0$, where $\mathcal{X}$ is the data space and $\mathbb{F}^u$ is the normalized high dimensional feature space. We also have access to a large unlabeled data pool $\mathcal{P}^u = \{x_i\}_{i\in[N]} \sim p_u$ inside data space $\mathcal{X}$ with distribution $p_u$, where $[N] = \{1, 2, \cdots, N\}$. 
We design a sampling strategy ${\mathcal{S}} = \{s_j \in [N]\}_{j\in[B]}$ to select a subset $\mathcal{P}^u_{\mathcal{S}} = \{x_{s_j}\}_{j\in[B]} \subset \mathcal{P}^u$ from $\mathcal{P}^u$, where $B$ is the annotation budget size for supervised finetuning. The model would have access to the labels $\{y_{s_j}\}_{j\in[B]} \subset \mathcal{Y}$ of this subset through the oracle, obtaining a labeled data pool $\mathcal{P}^l_{\mathcal{S}} = \{x_{s_j} , y_{s_j}\}_{j\in[B]}$, where $\mathcal{Y}$ is the label space.
The normalized high dimensional feature pool $\mathcal{F}^u = \{f_i\}_{i\in[N]} \sim p_{f_u}$ has a distribution $p_{f_u}$.
The feature pool $\mathcal{F}^u_{\mathcal{S}}$ is also associated with the selected data subset $\mathcal{P}^u_{\mathcal{S}}$, with the corresponding distribution over $\mathcal{F}^u_{\mathcal{S}}$ in the feature space denoted as $p_{f_{\mathcal{S}}}$ .

\subsection{Data Selection}
\label{sec:method:dist_sele}
The data selection strategy we employed is ActiveFT~\cite{ActiveFT_CVPR2023}, guided by two basic intuitions: 1) bringing the distributions of the selected subset $\mathcal{P}^u_{\mathcal{S}}$ and the original pool $\mathcal{P}^u \sim p_u$ closer, and 2) maintaining the diversity of $\mathcal{P}^u_{\mathcal{S}}$. The first ensures the model finetuned on the subset performs similarly to one trained on the full set, while the second allows the subset to cover corner cases in the full set.
The goal of distribution selection is to find the optimal selection strategy $\mathcal{S}$ as:
\begin{equation}
    \label{eq:DS_discrete}
    \mathcal{S}_{opt} = \underset{\mathcal{S}} {\arg\min} D(p_{f_u},p_{f_\mathcal{S}}) - \lambda R(\mathcal{F}^u_S)
\end{equation}
where $D(\cdot,\cdot)$ is a distance metric between distributions, $R(\cdot)$ is used to assess the diversity of a set, and $\lambda$ is a scaling factor to balance these two terms.

Optimizing the discrete selection strategy $\mathcal{S}$ directly is challenging. Therefore, it is better to model $p_{f_\mathcal{S}}$ with $p_{\theta_\mathcal{S}}$, where $\theta_\mathcal{S} = \{\theta^j_\mathcal{S}\}_{j\in[B]}$ are the continuous parameters and $B$ is the annotation budget size. Each $\theta^j_\mathcal{S}$ after optimization corresponds to the feature of a selected sample $f_{\mathcal{S}_j}$. The feature $f_{\mathcal{S}_j}$ closest to each $\theta^j_\mathcal{S}$ can be found after optimization to determine the selection strategy $\mathcal{S}$. Therefore, the goal in \cref{eq:DS_discrete} can be expressed as:
\begin{equation}
    \label{eq:DS_continuous}
    \theta_{\mathcal{S},opt} = \underset{\theta_\mathcal{S}} {\arg\min} D(p_{f_u},p_{\theta_\mathcal{S}}) - \lambda R(\theta_\mathcal{S}) ~~ s.t. ~~ {\left \| \theta^j_\mathcal{S} \right \|}_2 \! = \! 1
\end{equation}
The protocol outlined in~\cite{DINO_ICCV2021} is followed, and the cosine similarity between normalized features is utilized as the metric, denoted as $cos(f_1, f_2) = f_1^T f_2, {\left \| f_1 \right \|}_2 = {\left \| f_2 \right \|}_2 = 1$.
For each $f_i \in \mathcal{F}^u$, there exists a $\theta^{c_i}_\mathcal{S}$ most similar (and closest) to $f_i$, i.e.
\begin{equation}
    \label{eq:max_similar}
    c_i = \underset{j \in [B]}{\arg\max}~cos(f_i, \theta^j_\mathcal{S})
\end{equation}
where $c_i$ is continuously updated in the optimization process.

Thus, the following loss function can be continuously optimized to address \cref{eq:DS_continuous}:
\begin{equation}
\label{eq:loss_detailed}
\begin{split}
L &= \underset{\theta_\mathcal{S}} {\arg\min} D(p_{f_u},p_{\theta_\mathcal{S}}) - \lambda \cdot R(\theta_\mathcal{S}) \\
&= \! - \!\!\! \underset{f_i \in \mathcal{F}^u}{E} \!\! \left [ cos(f_i,\theta^{c_i}_\mathcal{S})/\tau \right ] \! + \!\!\! \underset{j \in [B]}{E} \!\!\! \left [ \! \log_{}{} 
\!\!\!\!\!\!\! \sum_{k \ne j,k \in [B]}^{} \!\!\!\!\!\!\! \exp(cos(\theta^{j}_\mathcal{S},\theta^{k}_\mathcal{S})/\tau ) \! \right] 
\end{split}
\end{equation}
where the balance weight $\lambda$ is empirically set to $1$, while the temperature scale $\tau$ is set to $0.07$.

Finally, the loss function in \cref{eq:loss_detailed} is directly optimized using gradient descent. After completing the optimization, the feature $\{ f_{s_j} \}_{j \in [B]}$ with the highest similarity to $\theta^j_\mathcal{S}$ is identified. 
The corresponding data samples $\{\mathrm{x}_{s_j} \}_{j \in [B]}$ are selected as the subset $\mathcal{P}^u_\mathcal{S}$ with selection strategy $\mathcal{S} = \{s_j \}_{j \in [B]}$.

\begin{figure}[!t]
\centering
\includegraphics[width=1.0\linewidth]{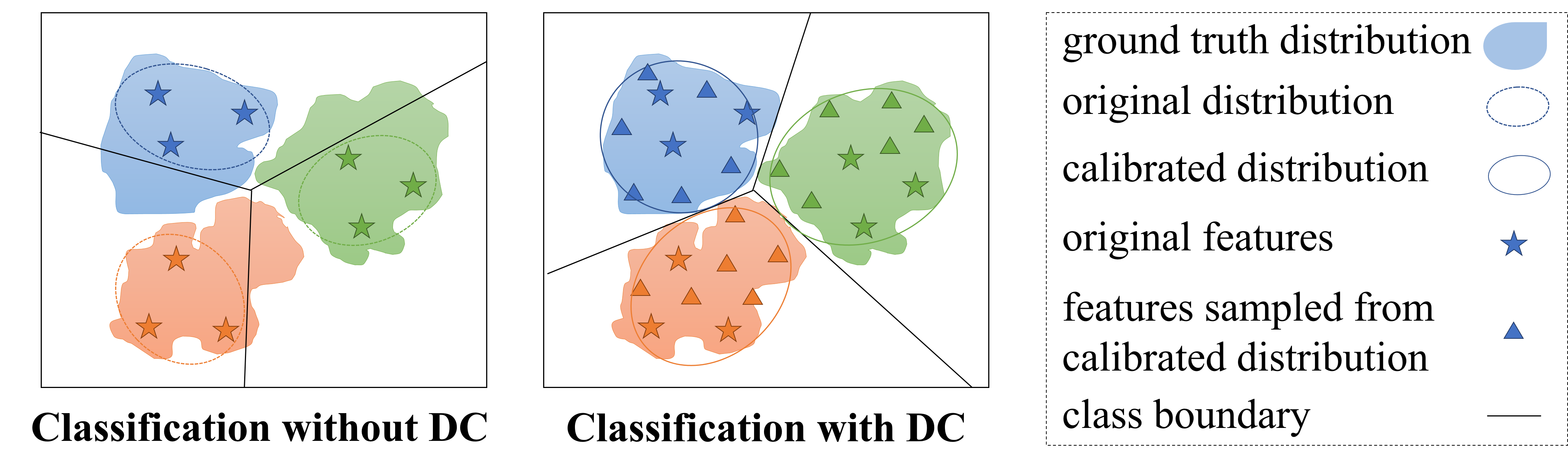}
\caption{
The diagram showcases two contrasting scenarios in the finetuning of a pretrained model. 
On the left, finetuning with a limited number of sample features leads to model overfitting. 
On the right, employing features sampled from a calibration distribution for finetuning the pretrained model demonstrates improved generalization.
}
\label{fig:DC_demo}
\end{figure}

\subsection{Distribution Calibration}
\label{sec:method:dist_cali}
Due to the typically limited number of labeled samples selected within the pretraining-finetuning paradigm, a substantial distributional discrepancy arises between the chosen subset and the entire data pool. This scenario, if unaddressed, may result in the overfitting of model parameters during the finetuning process, ultimately diminishing overall performance. 
In this section, we will not only use the labeled samples from \cref{sec:method:dist_sele} but also leverage information about the number of classification categories and other important information contained in a large number of unlabeled pretrained features.
To mitigate distributional bias, our method involves the selection of pseudo-labeled data that exhibit reliability and approximate the overarching distribution accurately, as shown in \cref{fig:DC_demo}.

\textbf{Tukey's Ladder of Powers Transformation}. 
To make the feature distribution more Gaussian-like, we first transform the features in $\mathcal{F}^u$ using Tukey's Ladder of Powers transformation~\cite{Log-Tukey_ICCVW2003}. 
Tukey's Ladder of Powers transformation belongs to a family of power transformations known for reducing distribution skewness and making distributions more Gaussian-like. 
This step is a prerequisite for the subsequent generation of features aligned with calibrated statistics conforming to a Gaussian distribution.
Tukey's Ladder of Powers transformation is formulated as:
\begin{equation}
\label{eq:Tukey}
\hat{x} =
\begin{cases}
\mathit{x^\lambda} ,  & \text{if $\lambda \ne 0$} \\
\mathit{\log x}, & \text{if $\lambda = 0$}
\end{cases}
\end{equation}
where $\lambda$ is a hyper-parameter used to control the distribution correction. 
The original feature can be recovered by setting $\lambda$ as 1. Decreasing $\lambda$ makes the distribution less positively skewed and vice versa.

\textbf{Statistics of Pseudo-Category}. Our initial procedure entails clustering the normalized high dimensional features, referred to as $\mathcal{F}^u$, based on the true number of categories. Subsequently, pseudo-labels are assigned to each feature, drawing upon the pool of labeled features $\mathcal{F}^l_\mathcal{S}$ for reference, where $\mathcal{F}^l_{\mathcal{S}} = \{f_{s_j} , y_{s_j}\}_{j\in[B]}$. Thus we get the pool of pseudo-labeled features $\mathcal{F}^u_\mathcal{C} = \{f_{\mathcal{C}_j} , y_{\mathcal{C}_j}\}_{j\in[N]}$, where $\mathcal{C}$ is the cluster method. Assuming there are a total of $K$ categories in $\mathcal{F}^u_\mathcal{C}$, the feature pool for category $y_{\mathcal{C}_j}$ is represented as $\mathcal{F}^i_\mathcal{C}= \{f_{\mathcal{C}_j}\}_{y_{\mathcal{C}_j}=i, j\in[N], i\in[K]}$.
Most of the features corresponding to the same real label in $\mathcal{F}^l_{\mathcal{S}}$ reside in the same pseudo-category, correcting $i$ to the corresponding real label.
\begin{equation}
\label{eq:mean_std}
  \begin{split}
  \mathrm{E}_i &= \frac{ {\textstyle \sum_{k=1}^{\left | \mathcal{F}^i_\mathcal{C} \right | } f_k} }{\left | \mathcal{F}^i_\mathcal{C} \right |} , ~~~ s.t. ~~ f_k \in \mathcal{F}^i_\mathcal{C}   \\
  \mathrm{S} ^2_i &=  \frac{1}{\left | \mathcal{F}^i_\mathcal{C} \right | - 1} \sum_{k=1}^{\left | \mathcal{F}^i_\mathcal{C} \right |} (f_k-\mathrm{E}_i)(f_k-\mathrm{E}_i)^T
  \end{split}   
\end{equation}
According to \cref{eq:mean_std}, we can get the mean and covariance of each pseudo-category feature pool.
A similar process can be employed to calculate the mean of features corresponding to category $i$ within $\mathcal{F}^l_\mathcal{S}$, denoted as $\mu^l_i$.

\textbf{Statistics Calibration}. 
To eliminate the effect of outliers on the feature mean $\mathrm{E}_i$, we recalculate the mean of the features by taking $90\%$ of the features that are closest to the mean, denoted as $\hat{\mu_i}$.
The calibrated mean and covariance are denoted as $\mu _i$ and $\sigma^2_i$, respectively, 
formulated as:
\begin{equation}
\label{eq:mean_std_fixed}
  \mu _i = (1-\beta) \hat{\mu}_i + \beta \mu_i^l ~~,~~
  \sigma^2_i = \mathrm{S} ^2_i + \xi
\end{equation}
where $\xi$ is a hyper-parameter that determines the degree of dispersion of features sampled from the calibrated distribution.
The term $(1-\beta)$ represents the degree to which the pseudo-category center contributes to the central information of the sample. 
\begin{equation}
\label{eq:center_fixed}
  \beta = 1 - \mathrm{e}^{-\alpha r} , ~~~ s.t. ~~ r \in [0, 100]
\end{equation}
In \cref{eq:center_fixed}, $\alpha$ is a hyper-parameter whose value determines the parameter $\beta$ in \cref{eq:mean_std_fixed}. The parameter $\beta$ represents the degree to which the true labeled data acquired through data selection contributes to the central information of the sample. 
The sampling ratio is denoted as $r\%$. For instance, when $r$ equals 1, the sampling ratio is 1\%.
Based on our understanding of the data distribution and experimental findings, we set the hyperparameter $\alpha$ to 0.7, 0.07, and 0.14 for CIFAR10, CIFAR100, and ImageNet, respectively.

\textbf{Feature Generation and Filtering}. With a set of calibrated statistics $\mathbb{S}_y = \{ \mu_y, \sigma^2_y \}$ for class $y=i$ in a target task, we generate a set of feature vectors with label $y$ by sampling from the calibrated Gaussian distributions 
defined as:
\begin{equation}
\label{eq:Gaussian_distribution}
  \mathbb{G}_y = \{ (f^g_y, y) \mid f^g_y \sim \mathcal{N}(\mu_y, \sigma^2_y), \forall (\mu_y, \sigma^2_y) \in \mathbb{S}_y\}
\end{equation}
The process of feature generation relies on creating $n$-fold features within the same class, and this is dependent on the number of features and classes in the labeled feature pool $\mathcal{F}^l_\mathcal{S}$.
We identify the real features in the unlabeled pool that have the highest similarity to the generated features, formulated as:
\begin{equation}
\label{eq:real_features}
  \mathbb{R}_y = \{ (\hat{f}^g_y, y) \mid \max~cos(f^g_y, \hat{f}^g_y), \forall (f^g_y, y) \!\in\! \mathbb{G}_y, \exists \hat{f}^g_y \!\in\! \mathcal{F}^u \}
\end{equation}
The data samples $\hat{\mathbb{R}}_y$ corresponding to $\mathbb{R}_y$, in conjunction with the pool of labeled data samples $\mathcal{P}^l_\mathcal{S}$, constitute the extended labeled pool $\mathcal{P}_{EL}$. This pool is subsequently utilized as the finetuning dataset for the pretrained model. 

\begin{algorithm}
\caption{Pseudo-code for ActiveDC}
\label{alg:ActiveDC}
  \newcommand\mycommfont[1]{\footnotesize\ttfamily\textcolor{blue}{#1}}
  \SetCommentSty{mycommfont}
  \SetKwData{Fu}{$\mathcal{F}^u$}
  \SetKwData{Pe}{$\mathcal{P}_{EL}$}
  \SetKwData{Fs}{$\mathcal{F}^u_{\mathcal{S}}$}
  \SetKwData{Ps}{$\mathcal{P}^u_{\mathcal{S}}$}
  \SetKwData{Fsl}{$\mathcal{F}^l_{\mathcal{S}}$}
  \SetKwData{Psl}{$\mathcal{P}^l_{\mathcal{S}}$}
  \SetKwData{Fc}{$\{\mathcal{F}^i_{\mathcal{C}}\}_{i \in [K]}$}
  \SetKwData{MU}{$\mu_i$}
  \SetKwData{SGM}{$\sigma^2_i$}
  \SetKwData{Guass}{$\mathcal{N}(\MU_, \SGM)$}
  \SetKwData{Fg}{$\{f_{gene.}, y=i\}_{\times n}$}
  \SetKwData{Fr}{$\{f_{real}, y=i\}_{\times n}$}
  \SetKwData{Ry}{$\mathbb{R}_y$}
  \SetKwData{hatRy}{$\hat{\mathbb{R}}_y$}
  \SetKwFunction{ActiveFT}{$\mathrm{ActiveFT}$}
  \SetKwFunction{Label}{$\mathrm{oracleLabel}$}
  \SetKwFunction{Trans}{$\mathrm{Transform}$}
  \SetKwFunction{Clust}{$\mathrm{Cluster}$}
  \SetKwFunction{Calib}{$\mathrm{Calibrate}$}
  \SetKwFunction{Sim}{$\mathrm{Sim}$}
  \SetKwFunction{Filter}{$\mathrm{Filter}$}
  \SetKwInOut{Input}{input}\SetKwInOut{Output}{output}

  \Input{the unlabeled feature pool \Fu}
  \Output{the extended labeled pool \Pe}
  \BlankLine
  \tcp{\textit{Data Selection by optimizing \cref{eq:loss_detailed}}}
  \Fs, \Ps $\leftarrow$ \ActiveFT{\Fu}\;
  \tcp{\textit{Oracle label data selected}}
  \Fsl, \Psl $\leftarrow$ \Label{\Fs, \Ps}\;
  \tcp{\textit{Powers Transformation As Per \cref{eq:Tukey}}}
  \Fu $\leftarrow$ \Trans{\Fu}\;
  \tcp{\textit{Clustering via K-Means}}
  \Fc $\leftarrow$ \Clust{\Fu}\;
  \tcp{\textit{Calibrate Statistics by \cref{eq:mean_std,eq:mean_std_fixed}}}
  \MU, \SGM $\leftarrow$ \Calib{\Fc, \Fsl}\;
  \tcp{\textit{Initialize pseudo-label feature pool}}
  $\Ry_{, \forall y \in [K]} \leftarrow$ \{\}\;
  \For{$i \leftarrow 0$ \KwTo $K$}{
    \tcp{\textit{Feature Generation As Per \cref{eq:Gaussian_distribution}}}
    \Fg $\sim$ \Guass\;
    \tcp{\textit{Real feat. most similar per \cref{eq:real_features}}}
    \Fr $\leftarrow$ \Sim{\Fg}\;
    \tcp{\textit{Filter harmful features}}
    \Ry $\leftarrow$ \Ry $\bigcup$ \Filter{\Fr}\;
  }
  \tcp{\textit{Constitute the extended labeled pool}}
  \Pe $\leftarrow$ \Psl $\bigcup \hatRy_{, \forall y \in [K]}$\;
  
\end{algorithm}

Additionally, it is essential to highlight that the generated features undergo a filtering process. 
The feature pool consisting of all the features in $\mathbb{R}_y$ is denoted as $\mathcal{F}^u_{\mathcal{G}} = \{\hat{f}^g_y\}_{(\hat{f}^g_y, y) \in \mathbb{R}_y}$.
The extended feature pool $\mathcal{F}^u_E~( \mathcal{F}^u_\mathcal{S} \bigcup \mathcal{F}^u_\mathcal{G})$ is also associated with the extended data subset in $\mathcal{P}_{EL}$, with the corresponding distribution over $\mathcal{F}^u_{E}$ in the feature space denoted as $p_{f_{E}}$.
The Earth Mover's Distance (EMD) metric~\cite{EMD_ICCV1998} is employed as a quantitative measure for assessing the dissimilarity between a subset distribution $p_{f_E}$ and the overall distribution $p_{f_u}$. 
Its application enables the identification and elimination of generative features that pose detriment to the overall distribution.
More details about filtering process are available in the supplementary materials.
Our method is summarised in \cref{alg:ActiveDC}.

\section{Experiments}
\label{sec:exp}

Our method is evaluated on three image classification datasets of different scales with different sampling ratios. 
We compare the evaluation results with several baseline algorithms, traditional active learning algorithms, and ActiveFT, an efficient active finetuning algorithm. 
These will be presented in \cref{sec:exp:setting} and \cref{sec:exp:results}.
We provide both qualitative and quantitative analysis of our method in \cref{sec:exp:analysis}. 
Finally, we investigate the role of different modules and different values of hyperparameters in our method in \cref{sec:exp:ablation}. 
The experiments were conducted using two GeForce RTX 3090 (24GB) GPUs, employing the DistributedDataParallel technique to accelerate the finetuning process.

\subsection{Experiment Settings}
\label{sec:exp:setting}

\textbf{Dataset and Metric}. 
Our method, ActiveDC, is applied and evaluated on three well-established image classification datasets of varying classification scales. 
These datasets, namely CIFAR10, CIFAR100~\cite{cifar100_2009}, and ImageNet-1k~\cite{imagenet_IJCV2015}, each present distinct characteristics. 
CIFAR10 and CIFAR100 consist of 60,000 images, with 10 and 100 categories, respectively. Both datasets use 50,000 images for training and 10,000 for testing. In contrast, ImageNet-1k has 1,000 categories with 1,281,167 training images and 50,000 validation images.
The performance evaluation of our method is conducted using the \textit{Top-1 Accuracy} metric.

\textbf{Baselines}. 
We compare our method to five traditional active learning methods and four baseline methods, including the strong baseline ActiveFT, the first efficient method applied to active finetuning tasks. 
The five active learning algorithms, namely CoreSet~\cite{CoreSet_ICLR2018}, VAAL~\cite{VAAL_ICCV2019}, LearnLoss~\cite{LLoss_CVPR2019}, TA-VAAL~\cite{TA-VAAL_CVPR2021}, and ALFA-Mix~\cite{ALFA-Mix_CVPR2022}, have been extended and adapted to the active finetuning task. 
These selected active learning methods cover both diversity-based and uncertainty-based strategies within the active learning domain. 
The set of baseline methods comprises four distinct techniques, including random selection, K-Center-Greedy, K-Means, and ActiveFT algorithm~\cite{ActiveFT_CVPR2023}. 
These baseline methods serve as crucial reference points for the comprehensive evaluation of active finetuning approaches.

\noindent - \textbf{Random:} A straightforward baseline method involves the random selection of $B$ samples from the unlabeled pool, where $B$ is denoted as the annotation budget. \\
\noindent - \textbf{FDS:} $a.k.a$ K-Center-Greedy algorithm. This method entails the selection of the next sample feature that is the farthest from the current selections. 
It is designed to minimize the disparity between the expected loss of the entire pool and that of the selected subset. \\
\noindent - \textbf{K-Means:} 
In this context, the value of $K$ is set to the budget size, denoted as $B$. 
Overclustering is a strategy in which more clusters or subgroups are intentionally created than the expected number of distinct classes or groups in the data. 
Many studies in the unsupervised domain~\cite{SwAV_NIPS2020,IIC_ICCV2019,CrOC_CVPR2023} have shown that overclustering leads to better performance. \\
\noindent - \textbf{ActiveFT:} A method that constructs a representative subset from an unlabeled pool, aligning its distribution with the overall dataset while optimizing diversity through parametric model optimization in a continuous space.

\textbf{Implementation details}. 
In unsupervised pretraining phase, we adopt DeiT-Small architecture~\cite{DeiT_ICML2021} pretrained within the DINO framework~\cite{DINO_ICCV2021}, a well-established and effective choice on the ImageNet-1k dataset~\cite{imagenet_IJCV2015}. 
For consistency throughout the process, all images are resized to 224$\times$224.
In the data selection phase, the parameters denoted as $\theta_\mathcal{S}$ are optimized employing the Adam optimizer~\cite{Adam_2014} with a learning rate of $1\mathrm{e}{-3}$ until convergence. 
In the distribution calibration phase, the unsupervised clustering method and similarity retrieval mechanisms employed primarily rely on the FAISS 
(Facebook AI Similarity Search) library. 
Specifically, we employ the GPU-accelerated variant of K-Means for clustering and cosine similarity as the similarity metric. 
We use a full traversal search approach in FAISS for efficient retrieval. 
In the supervised finetuning phase, the DeiT-Small model follows the established protocol outlined in reference~\cite{DINO_ICCV2021}. 
The implementation of supervised finetuning is based on the official codebase of DeiT.
Further elaboration on the experiments is provided in the \textit{supplementary materials}.

\begin{table*}[!t]
\centering
\caption{
The experiments were conducted on different scale datasets with different sampling ratios. We report the average of multiple trials. 
The symbol ``-'' is used to indicate not applicable (N/A). Specifically, this symbol indicates cases where active learning cannot be applied because the sample size is too small.}
\resizebox{1.0\linewidth}{!}
{
\begin{tabular}{lccccccccccccc}
\hline
\multicolumn{1}{l|}{\multirow{2}{*}{\textbf{Methods}}}            & \multicolumn{5}{c|}{\textbf{CIFAR10}}                                                               & \multicolumn{4}{c|}{\textbf{CIFAR100}}                                              & \multicolumn{4}{c}{\textbf{ImageNet}} \\
                                             & \multicolumn{1}{|c}{0.1\%}         & 0.2\%         & 0.5\%         & 1\%           & \multicolumn{1}{c|}{2\%}           & 1\%           & 2\%           & 5\%           & \multicolumn{1}{c|}{10\%}          & 0.5\%          & 1\%          & 2\%               & 5\%               \\ \hline
\multicolumn{1}{l|}{\textbf{Random}}         & 36.7          & 49.3          & 77.3          & 82.2          & \multicolumn{1}{c|}{88.9}          & 14.9          & 24.3          & 50.8          & \multicolumn{1}{c|}{69.3}          & 29.9          & 45.1          & 53.0              & 64.3              \\
\multicolumn{1}{l|}{\textbf{FDS}}            & 27.6          & 31.2          & 64.5          & 73.2          & \multicolumn{1}{c|}{81.4}          & 8.1           & 12.8          & 16.9          & \multicolumn{1}{c|}{52.3}          & 19.9          & 26.7          & 42.3              & 55.5              \\
\multicolumn{1}{l|}{\textbf{K-Means}}        & 40.3          & 58.8          & 83.0          & 85.9          & \multicolumn{1}{c|}{89.6}          & 17.6          & 31.9          & 42.4          & \multicolumn{1}{c|}{70.7}          & 37.1          & 50.7          & 55.7                 & 62.2                 \\ \hline
\multicolumn{1}{l|}{\textbf{CoreSet}~\cite{CoreSet_ICLR2018}}        & -             & -             & -             & 81.6          & \multicolumn{1}{c|}{88.4}          & -             & 30.6          & 48.3          & \multicolumn{1}{c|}{62.9}          & -          & -          & -                 & 61.7              \\
\multicolumn{1}{l|}{\textbf{VAAL}~\cite{VAAL_ICCV2019}}           & -             & -             & -             & 80.9          & \multicolumn{1}{c|}{88.8}          & -             & 24.6          & 46.4          & \multicolumn{1}{c|}{70.1}          & -          & -          & -                 & 64.0              \\
\multicolumn{1}{l|}{\textbf{LearnLoss}~\cite{LLoss_CVPR2019}}      & -             & -             & -             & 81.6          & \multicolumn{1}{c|}{86.7}          & -             & 19.2          & 38.2          & \multicolumn{1}{c|}{65.7}          & -          & -          & -                 & 63.2              \\
\multicolumn{1}{l|}{\textbf{TA-VAAL}~\cite{TA-VAAL_CVPR2021}}        & -             & -             & -             & 82.6          & \multicolumn{1}{c|}{88.7}          & -             & 34.7          & 46.4          & \multicolumn{1}{c|}{66.8}          & -          & -          & -                 & 64.3              \\
\multicolumn{1}{l|}{\textbf{ALFA-Mix}~\cite{ALFA-Mix_CVPR2022}}       & -             & -             & -             & 83.4          & \multicolumn{1}{c|}{89.6}          & -             & 35.3          & 50.4          & \multicolumn{1}{c|}{69.9}          & -          & -          & -                 & 64.5              \\ \hline
\multicolumn{1}{l|}{\textbf{ActiveFT}~\cite{ActiveFT_CVPR2023}}       & 47.1          & 64.5          & 85.0          & 88.2          & \multicolumn{1}{c|}{90.1}          & 26.1          & 40.7          & 54.6          & \multicolumn{1}{c|}{71.0}          & 36.8          & 50.1          & 54.2              & 65.3              \\ \hline
\multicolumn{1}{l|}{\textbf{ActiveDC (ours)}} & \textbf{61.3} & \textbf{73.1} & \textbf{87.3} & \textbf{88.9} & \multicolumn{1}{c|}{\textbf{90.3}} & \textbf{34.5} & \textbf{54.6} & \textbf{71.9} & \multicolumn{1}{c|}{\textbf{74.3}} & \textbf{50.9}     & \textbf{56.3}     & \textbf{60.1}     & \textbf{68.2}     \\ \hline
\end{tabular}
}
\label{tab:main_result}
\end{table*}

\subsection{Overall Results}
\label{sec:exp:results}

Our reported results are derived from meticulous averaging across three independent experimental runs, presented in \cref{tab:main_result}. 
The traditional active learning methods tend to falter within the pretraining-finetuning paradigm, as documented in prior works~\cite{SSmeetAL_ICCV2021,ActiveFT_CVPR2023,TypiClust_ICML2022}.
In stark contrast, our method, ActiveDC, exhibits superior performance across all three datasets, even at varying sampling ratios. 
Particularly noteworthy is the substantial performance improvement observed with a lower sampling ratio. 
This improvement can be attributed to our method's capability to not only select the most representative samples but also effectively calibrate the distribution of the sampled data. 
This observed phenomenon carries practical significance, especially in scenarios where the number of samples utilized in the pretraining-finetuning paradigm is considerably smaller than the size of the available pool. Such efficiency gains contribute to substantial cost savings in terms of annotation expenditures.
For instance, our method demonstrates a significant accuracy increase of more than $10\%$ when applied to the CIFAR10 dataset with a sampling rate of less than $0.2\%$. 
Similarly, when applied to the CIFAR100 dataset with a sampling rate of less than $5\%$, our method exhibits a similar improvement. This performance is in comparison to the strong benchmark ActiveFT~\cite{ActiveFT_CVPR2023}.

\subsection{Analysis}
\label{sec:exp:analysis}

\textbf{Generality of our Method}.
Our Distributed Calibration Module has the inherent flexibility to seamlessly integrate with a variety of active finetuning selection strategies. 
As indicated in \cref{tab:generality_DC}, we have conducted experiments on datasets of varying scales. The findings demonstrate the efficacy of our method across diverse diversity-based selection strategies.

\begin{table}[!t]
\centering
\caption{Generality of Distributed Calibration Module.}
\resizebox{1.0\linewidth}{!}{%
\begin{tabular}{l|cccccc}
\hline
\multirow{2}{*}{\textbf{Methods}} & \multicolumn{2}{c}{CIFAR10 (0.5\%)} & \multicolumn{2}{c}{CIFAR100 (5\%)} & \multicolumn{2}{c}{ImageNet (1\%)} \\ \cline{2-7} 
                   & w.o.DC   & w.DC        & w.o.DC   & w.DC        & w.o.DC   & w.DC  \\ \hline
\textbf{Random}    & 77.3  & \textbf{86.2}  & 50.8  & \textbf{70.1}  & 45.1  & \textbf{54.6} \\
\textbf{FDS}       & 64.5  & \textbf{78.3}  & 16.9  & \textbf{33.4}  & 26.7  & \textbf{40.2} \\
\textbf{K-Means}   & 83.0  & \textbf{86.8}  & 42.4  & \textbf{66.8}  & 50.7  & \textbf{56.8} \\
\textbf{ActiveFT}  & 85.0  & \textbf{87.3}  & 54.6  & \textbf{71.9}  & 50.1  & \textbf{56.3} \\ \hline
\end{tabular}
}
\label{tab:generality_DC}
\end{table}

\textbf{Data Selection Efficiency}. 
Efficient data selection is crucial, requiring both time-efficient and effective methods. 
In \cref{tab:time_cost}, we compare the time needed to select different ratios of training samples from CIFAR100. 
Traditional active learning involves repeated model training and data sampling, with training being the major time-consuming factor. In contrast, active finetuning methods select all samples in a single pass, eliminating the need for iterative model retraining in the selection process.
Our method, ActiveDC, increases processing time slightly compared to ActiveFT due to clustering and subsequent similarity retrieval after feature generation. However, the additional time, measured in minutes, is negligible compared to the significant performance improvements.

\begin{table}[!t]
\centering
\caption{Data selection efficiency of different methods.}
\resizebox{0.85\linewidth}{!}{
\begin{tabular}{lcccccc}
\hline
\textbf{Methods}             & 2\%             & 5\%           & 10\%        \\ \hline
\textbf{CoreSet}             & 15h7m           & 7h44m         & 20h38m      \\
\textbf{VAAL}                & 7h52m           & 12h13m        & 36h24m      \\
\textbf{LearLoss}            & 20m             & 1h37m         & 9h09m       \\
\textbf{K-Means }            & 16.6s           & 37.0s         & 70.2s       \\
\textbf{ActiveFT }           & 12.6s           & 21.9s         & 37.3s       \\
\textbf{ActiveDC (ours)}     & 2m40s           & 4m30s         & 7m20s       \\ \hline
\end{tabular}
}
\label{tab:time_cost}
\vspace{-10pt}
\end{table}

\begin{figure}[!t]
\centering
\includegraphics[width=0.9\linewidth]{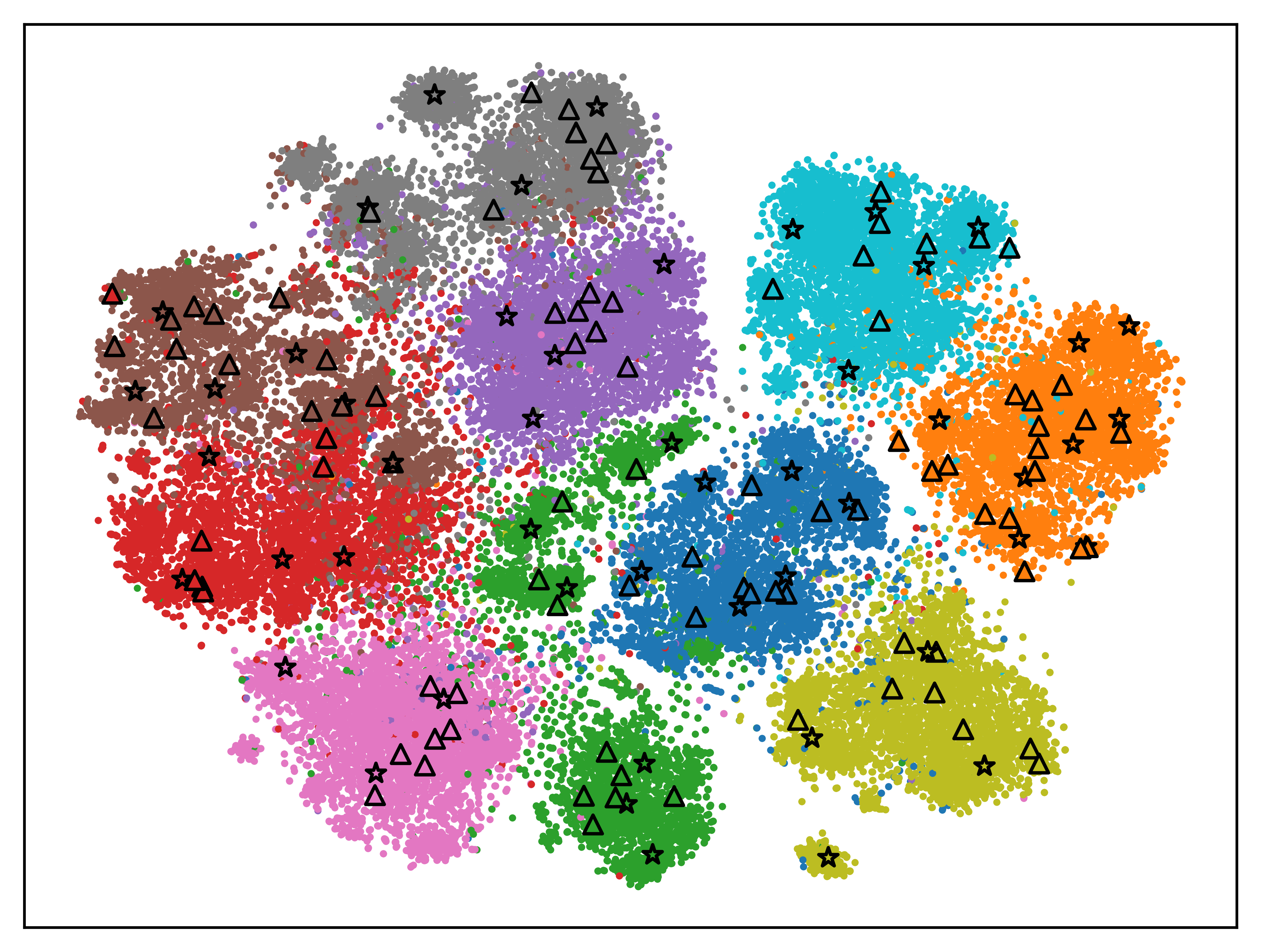}
\vspace{-10pt}
\caption{t-SNE Embeddings of CIFAR10: 
We visualize the embedding of selected samples labeled by the oracle (represented by a pentagram) and distribution calibration samples via ActiveDC (represented by a triangular shape) at a sampling ratio of $0.1$\%.
Best viewed in color.
}
\label{fig:tSNE_cifar10}
\vspace{-10pt}
\end{figure}

\textbf{Visualization of Selected Samples}. 
\Cref{fig:tSNE_cifar10} depicts the visualization of features extracted from the CIFAR10 training dataset. We utilize t-distributed Stochastic Neighbor Embedding (t-SNE)~\cite{tSNE_2008} for dimensionality reduction, employing distinct colors to differentiate feature categories. The pentagram symbolizes a $0.1\%$ sample selected based on its distributional similarity to the overall sample, while the triangle represents a pseudo-labeled sample chosen after the distribution calibration. This visualization offers an effective way to intuitively understand the impact of our method on sample selection in terms of distribution calibration. For a comprehensive understanding, we recommend consulting \cref{fig:DC_demo} in conjunction with this visualization.

\subsection{Ablation Study}
\label{sec:exp:ablation}
\begin{figure}[!t]
\centering
\includegraphics[width=0.9\linewidth]{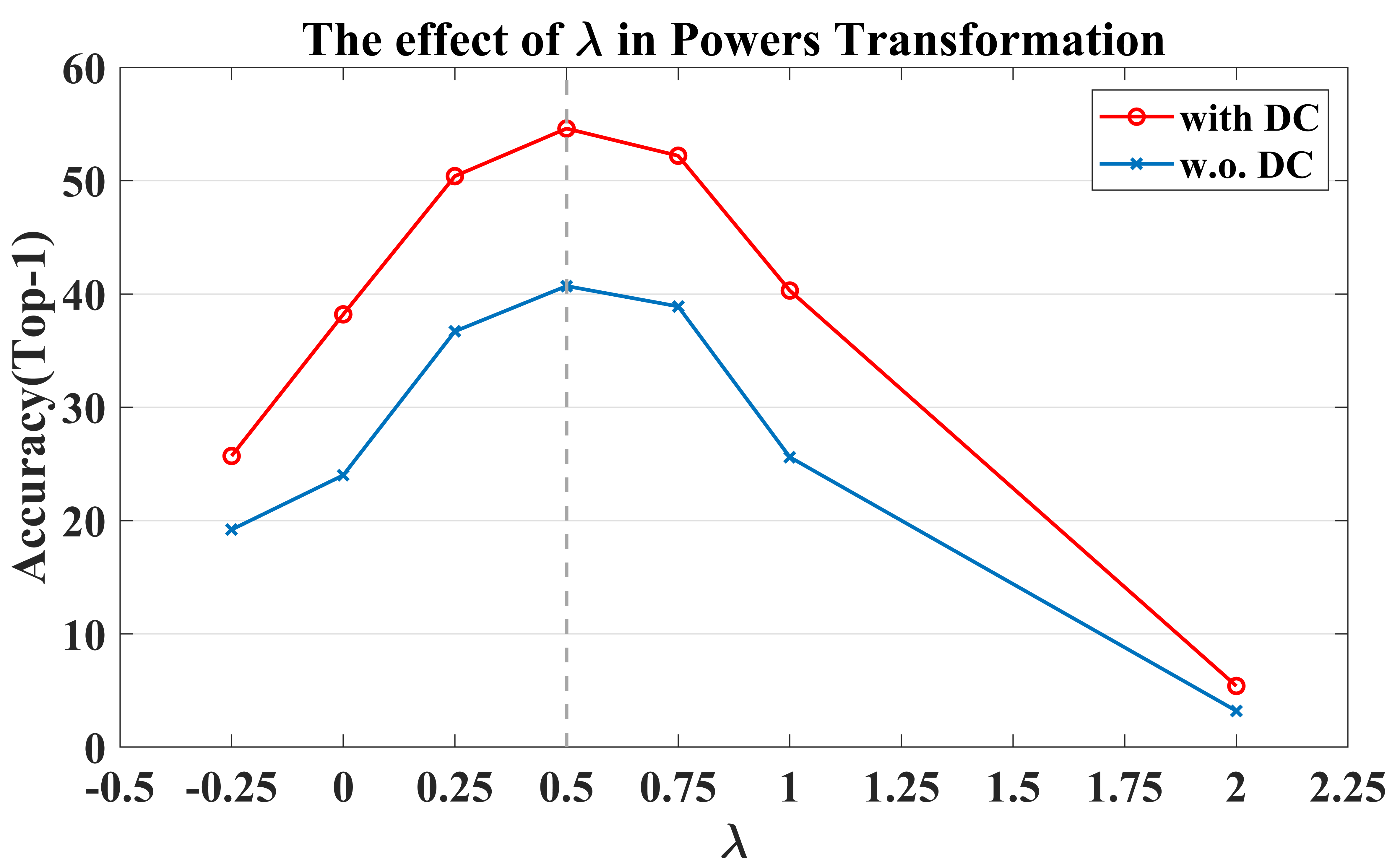}
\caption{The effect of $\lambda$:
The top fold (in red) represents finetuning accuracy with statistical calibration, while the lower fold (in blue) represents finetuning accuracy without statistical calibration. These results are obtained using different values of hyperparameter $\lambda$ in \cref{eq:Tukey}.
}
\label{fig:hyper_lambda}
\vspace{-10pt}
\end{figure}

\textbf{Choices of Power for Tukey's Transformation}.
In our analysis, we systematically investigate the impact of varying the hyperparameter $\lambda$ in \cref{eq:Tukey} on classification accuracy during Tukey's Ladder of Powers transformation process. Notably, the most favorable accuracy results were consistently achieved when setting $\lambda$ to $0.5$ across all three datasets. \Cref{fig:hyper_lambda} illustrates the accuracy outcomes when finetuning the model with a $1\%$ data sample extracted from the CIFAR100 dataset.

\textbf{Hyperparameter Tuning for Statistics Calibration}.
Based on our understanding of the data distribution and experimental findings, we set the hyperparameter $\alpha$ in \cref{eq:center_fixed} to 0.7, 0.07, and 0.14 for CIFAR10, CIFAR100, and ImageNet, respectively.
For details regarding the selection of $\alpha$ values in our experiments, please refer to the supplementary material. 
We analyze the effect of different values of the hyperparameter $\xi$ on the classification accuracy during the statistical calibration process, as shown in \cref{tab:hyper_xi}. 
The hyperparameter $\xi$ in \cref{eq:mean_std_fixed} determines the degree of dispersion of features sampled from the calibrated distribution.

\begin{table}[!t]
\centering
\caption{The effect of different values of $\xi$.}
\begin{tabular}{lccccc}
\hline
\textbf{hyperparameter $\xi$}  & \textbf{-0.1} & \textbf{0} & \textbf{0.1} & \textbf{0.2}  & \textbf{0.3} \\ \hline
CIFAR10 (0.5\%) & 85.5          & 86.4       & 86.9         & \textbf{87.3} & 70.2         \\
CIFAR100 (5\%)  & 65.3          & 68.6       & 71.5         & \textbf{71.9} & 61.8         \\
ImageNet (1\%)  & 48.7          & 50.6       & 56.0         & \textbf{56.3} & 50.2         \\ \hline
\end{tabular}
\label{tab:hyper_xi}
\vspace{-12pt}
\end{table}

\textbf{Number of Generated Features}. 
The number of generated features is twice the count of labeled features, as shown in \cref{tab:number_gene}. 
For each labeled feature, we generate two corresponding features, each adhering to the same calibrated distribution as the original feature. 
It is important to note that an excessive proliferation of pseudo-labeled data can lead to a decrease in accuracy. 
This phenomenon can be attributed to the presence of mislabeled data that is inadequately filtered out during the process described in \cref{sec:method:dist_cali}.

\begin{table}[!t]
\centering
\vspace{-10pt}
\caption{The effect of the number of generated features.}
\begin{tabular}{lcccc}
\hline
\textbf{number of gene.} & \textbf{$1\times$} & \textbf{$2\times$}   & \textbf{$3\times$} & \textbf{$4\times$} \\ \hline
CIFAR10 (0.5\%)          & 85.9        & \textbf{87.3} & 85.1        & 81.0        \\
CIFAR100 (5\%)           & 64.5        & \textbf{71.9} & 65.3        & 57.1        \\
ImageNet (1\%)           & 54.7        & \textbf{56.3} & 56.1        & 55.0        \\ \hline
\end{tabular}
\label{tab:number_gene}
\end{table}

\textbf{The effect of different module in DC}.
To investigate the influence of various sub-modules in distribution calibration on performance, we systematically conducted incremental experiments with these sub-modules, as detailed in \cref{tab:diff_modules_in_DC}. The rationale behind adopting an incremental manner lies in the interdependence of subsequent modules on their antecedent counterparts for functionality.
The experimental findings indicate that each submodule contributes positively to a certain extent.

\begin{table}[!t]
\centering
\vspace{-10pt}
\caption{The effect of different modules in Distributed Calibration.}
\resizebox{1.0\linewidth}{!}{%
\begin{tabular}{ccc|cc}
\hline
\multicolumn{3}{c|}{Distribution Calibration} & \multicolumn{2}{c}{CIFAR10}   \\ \hline
\makecell{Powers \\ Transformation} & \makecell{Statistics \\ Calibration} & \makecell{Feature \\ Filtering}                                          & 0.1\%         & 0.2\%   \\ \hline
\tiny{\XSolid}      & \tiny{\XSolid}        & \tiny{\XSolid}         & 47.1   & 64.5    \\
\tiny{\Checkmark}   & \tiny{\XSolid}        & \tiny{\XSolid}         & 50.3   & 66.8    \\
\tiny{\Checkmark}   & \tiny{\Checkmark}     & \tiny{\XSolid}         & 59.5   & 70.6    \\
\tiny{\Checkmark}   & \tiny{\Checkmark}     & \tiny{\Checkmark}      & \textbf{61.3} & \textbf{73.1} \\ \hline
\end{tabular}
}
\label{tab:diff_modules_in_DC}
\vspace{-15pt}
\end{table}

\section{Conclusion}
\label{sec:conc_and_limit}
This work proposes a novel method named ActiveDC, designed for data selection and distribution calibration of active finetuning tasks. The method consists of two crucial steps: Data Selection and Distribution Calibration. In the Distribution Calibration step, we leverage a substantial volume of unlabeled pretrained features to extract insights into class distribution and robustly calibrate the statistical information by ingeniously combining it with information from the labeled samples. Subsequently, we select pseudo-labeled data points that demonstrate reliability and provide an accurate approximation of the overall data distribution. 
Extensive experiments have demonstrated its effectiveness and significance. For future work, we will concentrate on optimizing efficiency, making it applicable to a broader range of application scenarios.

\noindent
\textbf{Acknowledgments.} 
This work was supported by the National Natural Science Foundation of China (No.~62302031, No.~62176017), and Zhejiang Provincial Natural Science Foundation of China (No.~LQ23F020024), 
and the Key R\&D Program of Zhejiang Province (No.~2024C01020).

\vspace{-20pt}
{
    \small
    \bibliographystyle{ieeenat_fullname}
    \bibliography{main}

\begin{thebibliography}{46}
\providecommand{\natexlab}[1]{#1}
\providecommand{\url}[1]{\texttt{#1}}
\expandafter\ifx\csname urlstyle\endcsname\relax
  \providecommand{\doi}[1]{doi: #1}\else
  \providecommand{\doi}{doi: \begingroup \urlstyle{rm}\Url}\fi

\bibitem[Agarwal et~al.(2020)Agarwal, Arora, Anand, and Arora]{CDAL_ECCV2020}
Sharat Agarwal, Himanshu Arora, Saket Anand, and Chetan Arora.
\newblock Contextual diversity for active learning.
\newblock In \emph{Computer Vision--ECCV 2020: 16th European Conference, Glasgow, UK, August 23--28, 2020, Proceedings, Part XVI 16}, pages 137--153. Springer, 2020.

\bibitem[Ash et~al.(2020)Ash, Zhang, Krishnamurthy, Langford, and Agarwal]{BADGE_ICLR2020}
Jordan~T. Ash, Chicheng Zhang, Akshay Krishnamurthy, John Langford, and Alekh Agarwal.
\newblock Deep batch active learning by diverse, uncertain gradient lower bounds.
\newblock In \emph{International Conference on Learning Representations}, 2020.

\bibitem[Bao et~al.(2022)Bao, Dong, Piao, and Wei]{BeiT_ICCV2022}
Hangbo Bao, Li Dong, Songhao Piao, and Furu Wei.
\newblock {BE}it: {BERT} pre-training of image transformers.
\newblock In \emph{International Conference on Learning Representations}, 2022.

\bibitem[Bengar et~al.(2021)Bengar, van~de Weijer, Twardowski, and Raducanu]{SSmeetAL_ICCV2021}
Javad~Zolfaghari Bengar, Joost van~de Weijer, Bartlomiej Twardowski, and Bogdan Raducanu.
\newblock Reducing label effort: Self-supervised meets active learning.
\newblock In \emph{Proceedings of the IEEE/CVF International Conference on Computer Vision}, pages 1631--1639, 2021.

\bibitem[Cabannes et~al.(2023)Cabannes, Bottou, Lecun, and Balestriero]{Active_SSL_ICCV2023}
Vivien Cabannes, Leon Bottou, Yann Lecun, and Randall Balestriero.
\newblock Active self-supervised learning: A few low-cost relationships are all you need.
\newblock In \emph{Proceedings of the IEEE/CVF International Conference on Computer Vision (ICCV)}, pages 16274--16283, 2023.

\bibitem[Caron et~al.(2020)Caron, Misra, Mairal, Goyal, Bojanowski, and Joulin]{SwAV_NIPS2020}
Mathilde Caron, Ishan Misra, Julien Mairal, Priya Goyal, Piotr Bojanowski, and Armand Joulin.
\newblock Unsupervised learning of visual features by contrasting cluster assignments.
\newblock \emph{Advances in neural information processing systems}, 33:\penalty0 9912--9924, 2020.

\bibitem[Caron et~al.(2021)Caron, Touvron, Misra, J{\'e}gou, Mairal, Bojanowski, and Joulin]{DINO_ICCV2021}
Mathilde Caron, Hugo Touvron, Ishan Misra, Herv{\'e} J{\'e}gou, Julien Mairal, Piotr Bojanowski, and Armand Joulin.
\newblock Emerging properties in self-supervised vision transformers.
\newblock In \emph{Proceedings of the IEEE/CVF international conference on computer vision}, pages 9650--9660, 2021.

\bibitem[Chen et~al.(2023)Chen, Bai, Huang, Lu, Wen, Yuille, and Zhou]{CSVAL_MIDL2023}
Liangyu Chen, Yutong Bai, Siyu Huang, Yongyi Lu, Bihan Wen, Alan Yuille, and Zongwei Zhou.
\newblock Making your first choice: To address cold start problem in medical active learning.
\newblock In \emph{Medical Imaging with Deep Learning}, 2023.

\bibitem[Chen et~al.(2020)Chen, Kornblith, Norouzi, and Hinton]{SimCLR_ICML2020}
Ting Chen, Simon Kornblith, Mohammad Norouzi, and Geoffrey Hinton.
\newblock A simple framework for contrastive learning of visual representations.
\newblock In \emph{International conference on machine learning}, pages 1597--1607. PMLR, 2020.

\bibitem[Chen and He(2021)]{SimSiam_CVPR2021}
Xinlei Chen and Kaiming He.
\newblock Exploring simple siamese representation learning.
\newblock In \emph{Proceedings of the IEEE/CVF conference on computer vision and pattern recognition}, pages 15750--15758, 2021.

\bibitem[Chen et~al.(2021)Chen, Xie, and He]{MoCov3_ICCV2021}
X Chen, S Xie, and K He.
\newblock An empirical study of training self-supervised vision transformers.
\newblock In \emph{CVF International Conference on Computer Vision (ICCV)}, pages 9620--9629, 2021.

\bibitem[Dosovitskiy et~al.(2021)Dosovitskiy, Beyer, Kolesnikov, Weissenborn, Zhai, Unterthiner, Dehghani, Minderer, Heigold, Gelly, Uszkoreit, and Houlsby]{ViT_ICLR2021}
Alexey Dosovitskiy, Lucas Beyer, Alexander Kolesnikov, Dirk Weissenborn, Xiaohua Zhai, Thomas Unterthiner, Mostafa Dehghani, Matthias Minderer, Georg Heigold, Sylvain Gelly, Jakob Uszkoreit, and Neil Houlsby.
\newblock An image is worth 16x16 words: Transformers for image recognition at scale.
\newblock In \emph{International Conference on Learning Representations}, 2021.

\bibitem[Farquhar et~al.(2021)Farquhar, Gal, and Rainforth]{ALB_ICLR2021}
Sebastian Farquhar, Yarin Gal, and Tom Rainforth.
\newblock On statistical bias in active learning: How and when to fix it.
\newblock In \emph{International Conference on Learning Representations}, 2021.

\bibitem[Ganatra(2023)]{Log-Tukey_ICCVW2003}
Vaibhav Ganatra.
\newblock Logarithm-transform aided gaussian sampling for few-shot learning.
\newblock In \emph{Proceedings of the IEEE/CVF International Conference on Computer Vision}, pages 247--252, 2023.

\bibitem[Grill et~al.(2020)Grill, Strub, Altch{\'e}, Tallec, Richemond, Buchatskaya, Doersch, Avila~Pires, Guo, Gheshlaghi~Azar, et~al.]{BYOL_NIPS2020}
Jean-Bastien Grill, Florian Strub, Florent Altch{\'e}, Corentin Tallec, Pierre Richemond, Elena Buchatskaya, Carl Doersch, Bernardo Avila~Pires, Zhaohan Guo, Mohammad Gheshlaghi~Azar, et~al.
\newblock Bootstrap your own latent-a new approach to self-supervised learning.
\newblock \emph{Advances in neural information processing systems}, 33:\penalty0 21271--21284, 2020.

\bibitem[Hacohen et~al.(2022)Hacohen, Dekel, and Weinshall]{TypiClust_ICML2022}
Guy Hacohen, Avihu Dekel, and Daphna Weinshall.
\newblock Active learning on a budget: Opposite strategies suit high and low budgets.
\newblock In \emph{International Conference on Machine Learning}, pages 8175--8195. PMLR, 2022.

\bibitem[He et~al.(2020)He, Fan, Wu, Xie, and Girshick]{MoCo_CVPR2020}
Kaiming He, Haoqi Fan, Yuxin Wu, Saining Xie, and Ross Girshick.
\newblock Momentum contrast for unsupervised visual representation learning.
\newblock In \emph{Proceedings of the IEEE/CVF conference on computer vision and pattern recognition}, pages 9729--9738, 2020.

\bibitem[He et~al.(2022)He, Chen, Xie, Li, Doll{\'a}r, and Girshick]{MAE_CVPR2022}
Kaiming He, Xinlei Chen, Saining Xie, Yanghao Li, Piotr Doll{\'a}r, and Ross Girshick.
\newblock Masked autoencoders are scalable vision learners.
\newblock In \emph{Proceedings of the IEEE/CVF conference on computer vision and pattern recognition}, pages 16000--16009, 2022.

\bibitem[Ji et~al.(2019)Ji, Henriques, and Vedaldi]{IIC_ICCV2019}
Xu Ji, Joao~F Henriques, and Andrea Vedaldi.
\newblock Invariant information clustering for unsupervised image classification and segmentation.
\newblock In \emph{Proceedings of the IEEE/CVF international conference on computer vision}, pages 9865--9874, 2019.

\bibitem[Joshi et~al.(2009)Joshi, Porikli, and Papanikolopoulos]{BvSB_CVPR2009}
Ajay~J Joshi, Fatih Porikli, and Nikolaos Papanikolopoulos.
\newblock Multi-class active learning for image classification.
\newblock In \emph{2009 ieee conference on computer vision and pattern recognition}, pages 2372--2379. IEEE, 2009.

\bibitem[Kim et~al.(2021)Kim, Park, Kim, and Chun]{TA-VAAL_CVPR2021}
Kwanyoung Kim, Dongwon Park, Kwang~In Kim, and Se~Young Chun.
\newblock Task-aware variational adversarial active learning.
\newblock In \emph{Proceedings of the IEEE/CVF Conference on Computer Vision and Pattern Recognition}, pages 8166--8175, 2021.

\bibitem[Kim et~al.(2023)Kim, Bae, Song, and Yun]{ReFAL_CVPR2023}
SangMook Kim, Sangmin Bae, Hwanjun Song, and Se-Young Yun.
\newblock Re-thinking federated active learning based on inter-class diversity.
\newblock In \emph{Proceedings of the IEEE/CVF Conference on Computer Vision and Pattern Recognition}, pages 3944--3953, 2023.

\bibitem[Kingma and Ba(2014)]{Adam_2014}
Diederik~P Kingma and Jimmy Ba.
\newblock Adam: A method for stochastic optimization.
\newblock \emph{arXiv preprint arXiv:1412.6980}, 2014.

\bibitem[Krizhevsky et~al.(2009)Krizhevsky, Hinton, et~al.]{cifar100_2009}
Alex Krizhevsky, Geoffrey Hinton, et~al.
\newblock Learning multiple layers of features from tiny images.
\newblock 2009.

\bibitem[Liu et~al.(2023)Liu, Yuan, Fu, Jiang, Hayashi, and Neubig]{liu2023pre}
Pengfei Liu, Weizhe Yuan, Jinlan Fu, Zhengbao Jiang, Hiroaki Hayashi, and Graham Neubig.
\newblock Pre-train, prompt, and predict: A systematic survey of prompting methods in natural language processing.
\newblock \emph{ACM Computing Surveys}, 55\penalty0 (9):\penalty0 1--35, 2023.

\bibitem[Lyu et~al.(2023)Lyu, Zhou, Chen, Huang, Yu, Li, Guo, Guo, Xiang, and Ding]{Box_Level_AD_CVPR2023}
Mengyao Lyu, Jundong Zhou, Hui Chen, Yijie Huang, Dongdong Yu, Yaqian Li, Yandong Guo, Yuchen Guo, Liuyu Xiang, and Guiguang Ding.
\newblock Box-level active detection.
\newblock In \emph{Proceedings of the IEEE/CVF Conference on Computer Vision and Pattern Recognition}, pages 23766--23775, 2023.

\bibitem[Parvaneh et~al.(2022)Parvaneh, Abbasnejad, Teney, Haffari, Van Den~Hengel, and Shi]{ALFA-Mix_CVPR2022}
Amin Parvaneh, Ehsan Abbasnejad, Damien Teney, Gholamreza~Reza Haffari, Anton Van Den~Hengel, and Javen~Qinfeng Shi.
\newblock Active learning by feature mixing.
\newblock In \emph{Proceedings of the IEEE/CVF Conference on Computer Vision and Pattern Recognition}, pages 12237--12246, 2022.

\bibitem[Rana and Rawat(2023)]{HAL_clustering_VAD_CVPR2023}
Aayush~J Rana and Yogesh~S Rawat.
\newblock Hybrid active learning via deep clustering for video action detection.
\newblock In \emph{Proceedings of the IEEE/CVF Conference on Computer Vision and Pattern Recognition}, pages 18867--18877, 2023.

\bibitem[Rubner et~al.(1998)Rubner, Tomasi, and Guibas]{EMD_ICCV1998}
Yossi Rubner, Carlo Tomasi, and Leonidas~J Guibas.
\newblock A metric for distributions with applications to image databases.
\newblock In \emph{Sixth international conference on computer vision (IEEE Cat. No. 98CH36271)}, pages 59--66. IEEE, 1998.

\bibitem[Russakovsky et~al.(2015)Russakovsky, Deng, Su, Krause, Satheesh, Ma, Huang, Karpathy, Khosla, Bernstein, et~al.]{imagenet_IJCV2015}
Olga Russakovsky, Jia Deng, Hao Su, Jonathan Krause, Sanjeev Satheesh, Sean Ma, Zhiheng Huang, Andrej Karpathy, Aditya Khosla, Michael Bernstein, et~al.
\newblock Imagenet large scale visual recognition challenge.
\newblock \emph{International journal of computer vision}, 115:\penalty0 211--252, 2015.

\bibitem[Sener and Savarese(2018)]{CoreSet_ICLR2018}
Ozan Sener and Silvio Savarese.
\newblock Active learning for convolutional neural networks: A core-set approach.
\newblock In \emph{International Conference on Learning Representations}, 2018.

\bibitem[Shui et~al.(2020)Shui, Zhou, Gagn{\'e}, and Wang]{WAAL_ICAIS2020}
Changjian Shui, Fan Zhou, Christian Gagn{\'e}, and Boyu Wang.
\newblock Deep active learning: Unified and principled method for query and training.
\newblock In \emph{International Conference on Artificial Intelligence and Statistics}, pages 1308--1318. PMLR, 2020.

\bibitem[Singh et~al.(2023)Singh, Duval, Alwala, Fan, Aggarwal, Adcock, Joulin, Doll{\'a}r, Feichtenhofer, Girshick, et~al.]{singh2023effectiveness}
Mannat Singh, Quentin Duval, Kalyan~Vasudev Alwala, Haoqi Fan, Vaibhav Aggarwal, Aaron Adcock, Armand Joulin, Piotr Doll{\'a}r, Christoph Feichtenhofer, Ross Girshick, et~al.
\newblock The effectiveness of mae pre-pretraining for billion-scale pretraining.
\newblock \emph{arXiv preprint arXiv:2303.13496}, 2023.

\bibitem[Sinha et~al.(2019)Sinha, Ebrahimi, and Darrell]{VAAL_ICCV2019}
Samarth Sinha, Sayna Ebrahimi, and Trevor Darrell.
\newblock Variational adversarial active learning.
\newblock In \emph{Proceedings of the IEEE/CVF International Conference on Computer Vision}, pages 5972--5981, 2019.

\bibitem[Stegm{\"u}ller et~al.(2023)Stegm{\"u}ller, Lebailly, Bozorgtabar, Tuytelaars, and Thiran]{CrOC_CVPR2023}
Thomas Stegm{\"u}ller, Tim Lebailly, Behzad Bozorgtabar, Tinne Tuytelaars, and Jean-Philippe Thiran.
\newblock Croc: Cross-view online clustering for dense visual representation learning.
\newblock In \emph{Proceedings of the IEEE/CVF Conference on Computer Vision and Pattern Recognition}, pages 7000--7009, 2023.

\bibitem[Touvron et~al.(2021)Touvron, Cord, Douze, Massa, Sablayrolles, and J{\'e}gou]{DeiT_ICML2021}
Hugo Touvron, Matthieu Cord, Matthijs Douze, Francisco Massa, Alexandre Sablayrolles, and Herv{\'e} J{\'e}gou.
\newblock Training data-efficient image transformers \& distillation through attention.
\newblock In \emph{International conference on machine learning}, pages 10347--10357. PMLR, 2021.

\bibitem[Tran et~al.(2019)Tran, Do, Reid, and Carneiro]{BGADL_ICML2019}
Toan Tran, Thanh-Toan Do, Ian Reid, and Gustavo Carneiro.
\newblock Bayesian generative active deep learning.
\newblock In \emph{International Conference on Machine Learning}, pages 6295--6304. PMLR, 2019.

\bibitem[Tukra et~al.(2023)Tukra, Hoffman, and Chatfield]{PerceptualMAE_CVPR2023}
Samyakh Tukra, Frederick Hoffman, and Ken Chatfield.
\newblock Improving visual representation learning through perceptual understanding.
\newblock In \emph{Proceedings of the IEEE/CVF Conference on Computer Vision and Pattern Recognition}, pages 14486--14495, 2023.

\bibitem[Van~der Maaten and Hinton(2008)]{tSNE_2008}
Laurens Van~der Maaten and Geoffrey Hinton.
\newblock Visualizing data using t-sne.
\newblock \emph{Journal of machine learning research}, 9\penalty0 (11), 2008.

\bibitem[Xie et~al.(2023{\natexlab{a}})Xie, Ding, Tomizuka, and Zhan]{FreeSel_NIPS2023}
Yichen Xie, Mingyu Ding, Masayoshi Tomizuka, and Wei Zhan.
\newblock Towards free data selection with general-purpose models.
\newblock In \emph{Thirty-seventh Conference on Neural Information Processing Systems}, 2023{\natexlab{a}}.

\bibitem[Xie et~al.(2023{\natexlab{b}})Xie, Lu, Yan, Yang, Tomizuka, and Zhan]{ActiveFT_CVPR2023}
Yichen Xie, Han Lu, Junchi Yan, Xiaokang Yang, Masayoshi Tomizuka, and Wei Zhan.
\newblock Active finetuning: Exploiting annotation budget in the pretraining-finetuning paradigm.
\newblock In \emph{Proceedings of the IEEE/CVF Conference on Computer Vision and Pattern Recognition}, pages 23715--23724, 2023{\natexlab{b}}.

\bibitem[Yang et~al.(2023)Yang, Zhao, Qi, Qiao, Shi, and Zhao]{ShrinkMatch_ICCV2023}
Lihe Yang, Zhen Zhao, Lei Qi, Yu Qiao, Yinghuan Shi, and Hengshuang Zhao.
\newblock Shrinking class space for enhanced certainty in semi-supervised learning.
\newblock In \emph{Proceedings of the IEEE/CVF International Conference on Computer Vision}, pages 16187--16196, 2023.

\bibitem[Yang et~al.(2021)Yang, Liu, and Xu]{FreeLunch_ICLR2021}
Shuo Yang, Lu Liu, and Min Xu.
\newblock Free lunch for few-shot learning: Distribution calibration.
\newblock In \emph{International Conference on Learning Representations}, 2021.

\bibitem[Yoo and Kweon(2019)]{LLoss_CVPR2019}
Donggeun Yoo and In~So Kweon.
\newblock Learning loss for active learning.
\newblock In \emph{Proceedings of the IEEE/CVF conference on computer vision and pattern recognition}, pages 93--102, 2019.

\bibitem[Zhou et~al.(2023)Zhou, Li, Qin, Liu, Pan, Zhang, Zhao, Gao, and Li]{SparseMAE_ICCV2023}
Aojun Zhou, Yang Li, Zipeng Qin, Jianbo Liu, Junting Pan, Renrui Zhang, Rui Zhao, Peng Gao, and Hongsheng Li.
\newblock Sparsemae: Sparse training meets masked autoencoders.
\newblock In \emph{Proceedings of the IEEE/CVF International Conference on Computer Vision}, pages 16176--16186, 2023.

\bibitem[Zhou et~al.(2022)Zhou, Wei, Wang, Shen, Xie, Yuille, and Kong]{iBOT_ICLR2022}
Jinghao Zhou, Chen Wei, Huiyu Wang, Wei Shen, Cihang Xie, Alan Yuille, and Tao Kong.
\newblock ibot: Image bert pre-training with online tokenizer.
\newblock \emph{International Conference on Learning Representations (ICLR)}, 2022.

\end{thebibliography}
}


\end{document}